\documentclass[sigconf]{acmart}

\AtBeginDocument{%
  \providecommand\BibTeX{{%
    \normalfont B\kern-0.5em{\scshape i\kern-0.25em b}\kern-0.8em\TeX}}}
    
\usepackage{algorithm}
\usepackage{algorithmic}
\usepackage[flushleft]{threeparttable}
\usepackage{amsmath}
\usepackage{multicol}
\usepackage{multirow}
\usepackage{xspace}
\usepackage{booktabs}
\usepackage{caption}
\usepackage{subcaption}
\newtheorem{remark}{Remark}
\newcommand{\method}{NGNN }
\newcommand{\stmethod}{NGNN}
\newcommand{\smethod}{ngnn }

\usepackage{xcolor}

\renewcommand\footnotetextcopyrightpermission[1]{} 
\setcopyright{none}
\begin{document}

\title{Network In Graph Neural Network}

\author{Xiang Song}
\email{xiangsx@amazon.com}
\affiliation{%
  \institution{AWS AI}
  \country{USA}
}

\author{Runjie Ma}
\email{runjie@amazon.com}
\affiliation{%
  \institution{AWS Shanghai AI Lab}
  \country{China}
}

\author{Jiahang Li}
\email{jiahanli@amazon.com}
\affiliation{%
  \institution{AWS Shanghai AI Lab}
  \country{China}
}

\author{Muhan Zhang}
\email{muhan@wustl.edu}
\affiliation{%
  \institution{Institute for Artificial Intelligence of Peking University}
  \country{China}
}

\author{David Paul Wipf}
\email{daviwipf@amazon.com}
\affiliation{%
  \institution{AWS Shanghai AI Lab}
  \country{China}
}

\renewcommand{\shortauthors}{Xiang and Runjie, et al.}

\begin{abstract}
Graph Neural Networks (GNNs) have shown success in learning from graph structured data containing node/edge feature information, with application to social networks, recommendation, fraud detection and knowledge graph reasoning. In this regard, various strategies have been proposed in the past to improve the expressiveness of GNNs. For example, one straightforward option is to simply increase the parameter size by either expanding the hidden dimension or increasing the number of GNN layers. However, wider hidden layers can easily lead to overfitting, and incrementally adding more GNN layers can potentially result in over-smoothing. 
In this paper, we present a model-agnostic methodology, namely Network In Graph Neural Network (\method), that allows arbitrary GNN models to increase their model capacity by making the model deeper. However, instead of adding or widening GNN layers, \method deepens a GNN model by inserting non-linear feedforward neural network layer(s) within each GNN layer. Although, some works mentioned that adding MLPs within GNN layers could benefit the performance, they did not systematically analyze the reason for the improvement, nor evaluate with numerous GNNs on large-scale graph datasets. In this paper, we demonstrate that \method can keep the model stable against either node feature or graph structure perturbations through an
analysis of it as applied to a GraphSage base GNN on ogbn-products data. Furthermore, we take a wide-ranging evaluation of \method on both node classification and link prediction tasks and show that \method works reliably across diverse GNN architectures. For instance, it improves the test accuracy of GraphSage on the ogbn-products by 1.6\% and improves the hits@100 score of SEAL on ogbl-ppa by 7.08\% and the hits@20 score of GraphSage+Edge-Attr on ogbl-ppi by 6.22\%. And at the time of this submission, it achieved two first places on the OGB link prediction leaderboard.

\end{abstract}



\maketitle

\section{Introduction}
Graph Neural Networks (GNNs) capture local graph structure and feature information in a trainable fashion to derive powerful node representations. They have shown promising success on multiple graph-based machine learning tasks ~\cite{ying2018graph,scarselli2008graph,hu2020open} and are widely adopted by various web applications including social network~\cite{ying2018graph,rossi2020temporal}, recommendation~\cite{berg2017graph,fan2019graph,yu2021self}, fraud detection~\cite{wang2019fdgars,li2019spam,liu2020alleviating}, etc.
Various strategies have been proposed to improve the expressiveness of GNNs~\cite{hamilton2017inductive,velivckovic2017graph,xu2018powerful,schlichtkrull2018modeling}. 

\begin{figure}[t]
    \centering
    \begin{subfigure}[b]{1.0\columnwidth}
        \centering
        \includegraphics[width=1.0\columnwidth]{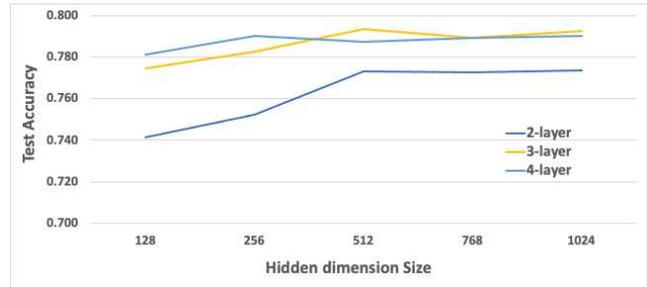}
        \caption{The test accuracy of GraphSage with different settings.}
        \label{fig:sage}
    \end{subfigure}
    \begin{subfigure}[b]{1.0\columnwidth}
        \centering
        \includegraphics[width=1.0\columnwidth]{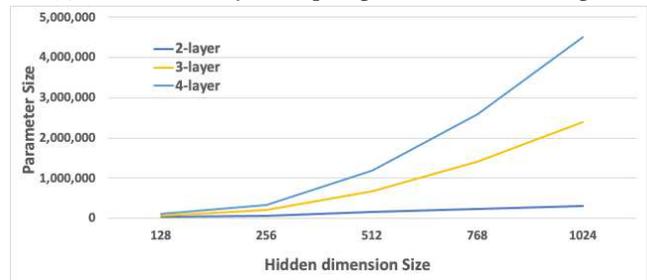}
        \caption{The model parameter sizes of GraphSage with different settings.}
        \label{fig:sage-zise}
    \end{subfigure}
\caption{The test accuracy of GraphSage on ogbn-products with different number of GNN layers (from 2 to 4) and different hidden dimension sizes (from 128 to 1024) and the corresponding model parameter sizes.}
\label{fig:product}
\end{figure}

One natural candidate for improving the performance of a GNN is to increase its parameter size by either expanding the hidden dimension or the number of GNN layers. However, this can result in a large computational cost with only a modest performance gain.  As a representative example, Figure~\ref{fig:product} displays the performance of GraphSage~\cite{hamilton2017inductive} under different settings on the ogbn-products dataset and the corresponding model parameter sizes. From these results, it can be seen that either increasing the hidden dimension or increasing the number of GNN layers increases the model parameter size exponentially, but brings little performance improvement in tersm of test accuracy. For example, in order to improve the accuracy of a 3-layer GraphSage model by 1\%, we need to add 2.3$\times$ more parameters (by increasing the hidden dimension from 256 to 512). Furthermore, with a larger hidden dimension a model is more likely to overfit the training data. On the other head, stacking multiple GNN layers may oversmooth the features of nodes~\cite{oono2019graph, chen2020measuring}. As shown in Figure~\ref{fig:sage}, GrageSage reaches its peak performance with only 3 GNN layers and a hidden dimension of 512.

Inspired by the Network-in-Network architecture~\cite{lin2013network}, we present Network-in-Graph Neural-Network (\method), a model agnostic methodology that allows arbitrary GNN models to increase  their model capacity by making the model deeper. However, instead of adding more GNN layers, \method deepens a GNN model by inserting non-linear feedforward neural network layer(s) within each GNN layer. This leads to a much smaller memory footprint than recent alternative deep GNN architectures~\cite{li2019deepgcns, li2020deepergcn} and can be applied to all kinds of GNN models with various training methods including full-graph training, neighbor sampling~\cite{hamilton2017inductive}, 
cluster-based sampling~\cite{chiang2019cluster} and local subgraph sampling~\cite{zhang2018link}. Thus, it can easily scale to large graphs.
Moreover, analysis of \method in conjunction with GraphSage on perturbed ogbn-products showed that \method is a cheap yet effective way to keep the model stable against either node feature or graph structure perturbations.

In this work, we applied \method to GCN~\cite{kipf2016semi}, GraphSage~\cite{hamilton2017inductive}, GAT~\cite{velivckovic2017graph} and AGDN~\cite{sun2020adaptive} and SEAL~\cite{zhang2018link}. We also combine the proposed technique with different mini-batch training methods including neighbor sampling, graph clustering and local subgraph sampling.
We conducted comprehensive experiments on several large-scale graph datasets for both node classification and link prediction leading to the following conclusions (which hold as of the time of this submission):
\begin{itemize}
    \item \method improves the performance of GraphSage and GAT and their variants on node classification datasets including ogbn-products, ogbn-arxiv, ogbn-proteins and reddit. It improves the test accuracy by 1.6\% on the ogbn-products datasets for GraphSage. Furthermore, \method with AGDN+BoT+self-KD+C\&S~\cite{huang2020combining} achieves the forth place on the ogbn-arxiv leaderboard\footnote{https://ogb.stanford.edu/docs/leader\_nodeprop/} and \method with GAT+BoT ~\cite{wang2021bag} achieves second place on the ogbn-proteins leaderboard with many fewer model parameters.
    \item \method improves the performance of SEAL, GCN and GraphSage and their variants on link prediction datasets including ogbl-collab, ogbl-ppa and ogbl-ppi. 
    For example, it increases the test hits@100 score by 7.08\% on the ogbl-ppa dataset for SEAL, which outperforms all the state-of-the-art approaches on the ogbl-ppa leaderboard\footnote{https://ogb.stanford.edu/docs/leader\_linkprop/} by a substantial margin. Furthermore, \method achieves an improvement of the test hits@20 score by 6.22\% on the ogbl-ppi dataset for GraphSage+EdgeAttr, which also takes the first place on the ogbl-ppi learderboard.
    \item \method improves the performance of GraphSage and GAT under different training methods including full-graph training, neighbor sampling, graph clustering, and subgraph sampling.
    \item \method is a more effective way of improving the model performance than expanding the hidden dimension. 
    It takes less parameter size and less training time to get better performance than simply doubling the hidden dimension.
\end{itemize}

In summary, we present \method, a method that deepens a GNN model without adding extra GNN message-passing layers. We show that \method significantly improves the performance of vanilla GNNs on various datasets for both node classification and link prediction. We demonstrate the generality of \method by applying them to various GNN architectures. 

\section{Related Work}
Deep models have been widely studied in various domains including computer vision~\cite{simonyan2014very,he2016deep}, natural language processing~\cite{brown2020language}, and speech recognition~\cite{zhang2017very}.
VGG~\cite{simonyan2014very} investigates the effect of the convolutional neural network depth on its accuracy in the large-scale image recognition setting. It demonstrates the depth of representations is essential to the model performance. But when the depth grows, the accuracy will not always grow. Resnet~\cite{he2016deep} eases the difficulties on training the deep model by introducing residual connections between input and output
layers. DenseNet~\cite{huang2017cvpr} takes this idea a step further by adding connections across layers. GPT-3~\cite{brown2020language} presents an autoregressive language model with 96 layers that achieves SOTA performance on various NLP tasks. Even so, while deep neural networks have achieved great success in various domains, the use of deep models in graph representation leaning is less well-established.

Most recent works~\cite{li2019deepgcns,li2020deepergcn,li2021training} attempt to train deep GNN models with a large number of parameters and achieved SOTA performance. For example,
DeepGCN~\cite{li2019deepgcns} adapts the concept of residual connections, dense connections, and dilated
convolutions~\cite{yu2015multi} to training very deep GCNs. However DeepGCN and its successor DeeperGCN~\cite{li2020deepergcn} have large memory footprints during model training which can be subject to current hardware limitations. RevGNN~\cite{li2021training} explored grouped reversible graph connections to train a deep GNN and has a much smaller memory footprint.
However, RevGNN can only work with full-graph training and cluster-based mini-batch training, which makes it difficult to work with other methods designed for large scale graphs such as neighbor sampling~\cite{hamilton2017inductive} and layer-wise sampling~\cite{chen2018fastgcn}. In contrast, \method deepens a GNN model by inserting non-linear feedforward layer(s) within each GNN layer. It can be applied to all kinds of GNN models with various training methods including full-graph training, neighbor sampling~\cite{hamilton2017inductive}, layer-wise sampling~\cite{chen2018fastgcn} and cluster-based sampling~\cite{chiang2019cluster}. 

Xu et al.~\cite{xu2018powerful} used Multilayer Perceptrons (MLPs) to learn the injective functions of the Graph Isomophism Network (GIN) model and showed its effectiveness on graph classification tasks. But they did not show whether adding an MLP within GNN layers works effectively across wide-ranging node classification and link prediction tasks.
Additionally, You et al.~\cite{you2020design} mentioned that adding MLPs within GNN layer could benefit the performance. However, they did not systematically analyze the reason for the performance improvement introduced by extra non-linear layers, nor evaluate with numerous SOTA GNN architectures on large-scale graph datasets for both node classification and link prediction tasks.  



\begin{figure}[t]
    \centering
    \begin{subfigure}[b]{1.0\columnwidth}
        \centering
        \includegraphics[width=1.0\columnwidth]{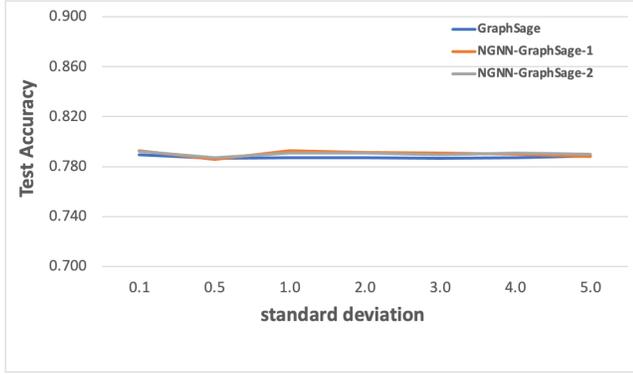}
        \caption{Gaussian noise is concatenated with node features.}
        \label{fig:cat-noise}
    \end{subfigure}
    \begin{subfigure}[b]{1.0\columnwidth}
        \centering
        \includegraphics[width=1.0\columnwidth]{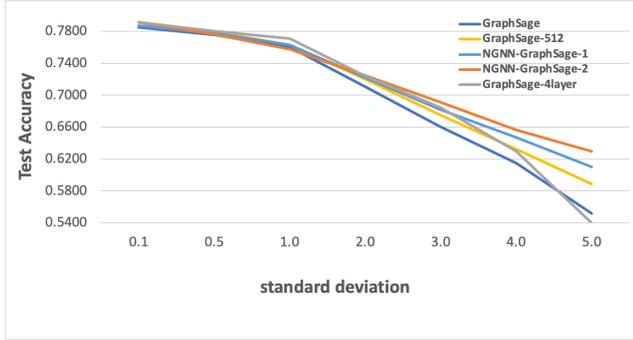}
        \caption{Gaussian noise is added to node features.}
        \label{fig:add-noise}
    \end{subfigure}
\caption{The test accuracy (\%) of GraphSage, GraphSage with the hidden dimension of 512 (GraphSage-512), 4-layer GraphSage (GraphSage-4layer), GraphSage with one additional non-linear layer in each GNN layer (\stmethod-GraphSage-1) and GraphSage with two additional non-linear layers in each GNN layer (\stmethod-GraphSage-2) on ogbn-product with randomly added Gaussian noise in node features. By default, a GraphSage model has three GNN layers with a hidden dimension of 256.}
\label{fig:noise}
\end{figure}

\section{Building Network in Graph Neural Network Models}
\subsection{Preliminaries}
A graph is composed of nodes and edges $\mathcal{G} = (\mathcal{V}, \mathcal{E})$ , where $\mathcal{V}=(v_1, \dots, v_N)$ is the set of $N$ nodes and $\mathcal{E} \subseteq \mathcal{V} \times \mathcal{V}$ is the set of edges.  Furthermore, $\mathcal{A} \in \{0, 1\}^{N\times N}$ denotes the corresponding adjacency matrix of $\mathcal{G}$. Let $\mathcal{X} \in \mathcal{R}^{N \times D}$ be the node feature space such that $\mathcal{X}=(x_1, \dots, x_N)^T$ where $x_i$ represents the node feature of $v_i$. Formally, the $(l+1)$-th layer of a GNN is defined as:\footnote{We omit edge features for simplicity.}

\begin{equation} \label{eq:1}
    h^{(l+1)} = \sigma(f_w(\mathcal{G}, h^l)),
\end{equation}

\noindent where the function $f_w(\mathcal{G}, h^l)$ is determined by learnable parameters $w$ and $\sigma(\cdot)$ is an optional activation function. Additionally, $h^l$ represents the embeddings of the nodes in the $l$-th layer, and $h^l = X$ when $l=1$. With an $L$-layer GNN, the node embeddings in the last layer $h^L$ are used by downstream tasks like node classification and link prediction. 

\subsection{Basic \method Design}
Inspired by the network-in-network architecture~\cite{lin2013network}, we deepen a GNN model by inserting non-linear feedforward neural network layer(s) within each GNN layer. The $(l+1)$-th layer in \method is thus constructed as:

\begin{equation} \label{eq:2}
    h^{(l+1)} = \sigma(g_{\smethod}(f_w(\mathcal{G}, h^l))).
\end{equation}

The calculation of $g_{\smethod}$ is defined layer-wise as:

\begin{equation} \label{eq:2}
    \begin{split}
    g_{\smethod}^1 = \sigma(f_w(\mathcal{G}, h^l)w^1) \\
    \dots \\
    g_{\smethod}^k = \sigma(g_{\smethod}^{k-1}w^k) \\
    \end{split}
\end{equation}

\noindent where $w_1, \dots, w_k$ are learnable weight matrices, $\sigma(\cdot)$ is an activation function, and $k$ is the number of in-GNN non-linear feedforward neural network layers. The first in-GNN layer takes the output of $f_w$ as input and performs the non-linear transformation.

\subsection{Discussion}
In this section, we demonstrate that a \method architecture can better handle both noisy node features and noisy graph structures relative to its vanilla GNN counterpart.

\begin{figure}[t]
    \centering
    \includegraphics[width=1.0\columnwidth]{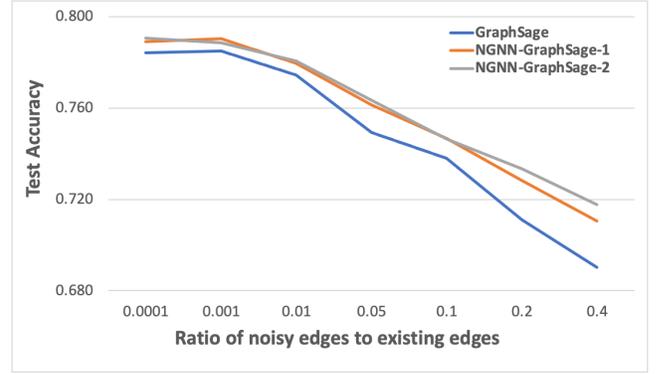}
\caption{The model performance of GraphSage, GraphSage with one additional non-linear layer in each GNN layer (\stmethod-GraphSage-1) and GraphSage with two additional non-linear layers in each GNN layer (\stmethod-GraphSage-2) on ogbn-product with randomly added noise edges.}
\label{fig:noise-edge}
\end{figure}

\begin{remark}
GNNs work well when the input features consist of distinguishable true features and noise. But when the true features are mixed with noise, GNNs can struggle to filter out the noise, especially as the noise level increases.
\end{remark}

GNNs follow a neural message passing scheme~\cite{gilmer2017neural} to aggregate information from neighbors of a target node. In doing so, they can perform noise filtering and learn from the resulting signal when 
the noise is in some way distinguishable from true features, such as when the latter are mostly low-frequency~\cite{nt2019revisiting}. However, when the noise level becomes too large and is mixed with true features, it cannot easily be reduced by GNNs~\cite{huang2019residual}. Figure~\ref{fig:noise} demonstrates this scenario.
Here we randomly added Gaussian noise $\mathcal{N} = N(0, \sigma)$ to node features in ogbn-products data, where $\sigma$ is the standard deviation ranging from 0.1 to 5.0.
We adopt two different methods for adding noise: 1) $\mathcal{X} = [\mathcal{X} | \mathcal{N}]$ as shown in Figure~\ref{fig:cat-noise}, where $|$ is a concatenation operation, and 2) $\mathcal{X} = [\mathcal{X} + \mathcal{N}]$ as shown in Figure~\ref{fig:add-noise}.
We trained GraphSage models (using the DGL ~\cite{wang2020deep} implementation) under five different settings: 1) baseline GraphSage with default 3-layer structure and hidden dimension of 256, denoted as GraphSage; 2) GraphSage with the hidden dimension increased to 512, denoted as GraphSage-512; 3) 4-layer GraphSage, denoted as GraphSage-4layer; 4) GraphSage with one additional non-linear layer in each GNN layer, denoted as \stmethod-GraphSage-1 and 5) GraphSage with two additional non-linear layers in each GNN layer, denoted as \stmethod-GraphSage-2. In all cases we used ReLU as the activation function.
As shown in Figure~\ref{fig:cat-noise}, GraphSage performs well when the noise is highly distinguishable from the true features. But the performance starts dropping when the noise is mixed with the true features and decays faster when $\sigma$ becomes larger than 1.0 as shown in Figure~\ref{fig:add-noise}.

The same scenario happens with the gfNN model~\cite{nt2019revisiting}, which is formed by transforming input node features via muliplications of the adjacency matrix followed by application of a single MLP block.
This relatively simple model was shown to be more noise tolerant than GCN and SGC~\cite{wu2019simplifying}; however, the performance of gfNN turns out to be much lower than the baseline GraphSage model in our experiments, so we do not present results here.

\begin{remark}
\method is a cheap yet effective way to form a GNN architecture that is stable against node feature perturbations.
\end{remark}

One potential way to improve the denoising capability of a GNN model is to increase the parameter count via a larger hidden dimension. As shown in Figure~\ref{fig:add-noise}, GraphSage-512 does perform better than the baseline GraphSage. But it is also more expensive as its parameter size (675,887) is $3.27\times$ larger than that of baseline GraphSage (206,895). And it is still not as effective as either \method model, both of which use considerably fewer parameters (see below) and yet have more stable performance as the noise level increases.

An alternative strategy for increasing the model parameter count is to add more GNN layers. As shown in Figure~\ref{fig:add-noise}, by adding one more GNN layer, GraphSage-4layer does outperform baseline GraphSage when $\sigma$ is smaller than 4.0. However, as a deeper GNN potentially aggregates more noisy information from its $L$ hop neighbors~\cite{zeng2020deep}, the performance of GraphSage-4layer drops below baseline GraphSage when $\sigma$ is 5.0.


In contrast to the above two methods, \stmethod-GraphSage achieves much better performance as shown in Figure~\ref{fig:add-noise} with fewer parameters (272,687 for \stmethod-GraphSage-1 and 338,479 for \stmethod-GraphSage-2) than GraphSage-512 and without introducing new GNN layers. It can help maintain model performance when $\sigma$ is smaller than 1.0 and slow the downward trend when $\sigma$ is larger than 1.0 compared to the other three counterparts. 

\begin{remark}
\method with GNNs can also keep the model stable against graph structure perturbation.
\end{remark}
We now show that by applying \method to a GNN, it can better deal with graph structure data perturbation. For this purpose, we randomly added $K$ edges to the original graph of ogbn-products, where $K$ is the ratio of newly added noise edges to the existing edges. For example $K=0.01$ means we randomly added 618.6K edges.\footnote{The graph of ogbn-product has 61,859,140 edges.}
We trained 3-layer GraphSage models with a hidden dimension of 256 under three different settings: 1) vanilla GraphSage, denoted as GraphSage; 2) GraphSage with one additional non-linear layer in each GNN layer, denoted as \method-GraphSage-1 and 3) GraphSage with two additional non-linear layers in each GNN layer, denoted as \method-GraphSage-2. Figure~\ref{fig:noise-edge} shows the results. It can be seen that, \method can help preserve the model performance when $K$ is smaller than 0.01 and ease the trend of performance downgrade after $K$ is larger than 0.01 comparing to vanilla GraphSage.

\section{Experiments}
\label{experiment}
We next provide experimental evidence to show that \method works will with various GNN architectures for both node classification and link prediction tasks in Sections~\ref{eval-nc} and  \ref{eval-lp}.
We also show that \method works with different training methods in Section~\ref{eval-train}. 
Finally, we discuss the impact of different \method settings in Sections~\ref{eval-mtl} and \ref{eval-dgnn}.

\subsection{Evaluation Setup}
\paragraph{Datasets} We conducted experiments on seven datasets, including ogbn-products, ogbn-arxiv and ogbn-proteins from ogbn~\cite{hu2020open} and reddit\footnote{http://snap.stanford.edu/graphsage/} 
for node classification, and ogbl-collab, ogbl-ppa and ogbl-citaiton2 from ogbl~\cite{hu2020open} for link prediction. The detailed statistics are summarized in Table~\ref{tab:dataset}.

\begin{table}[t]
\centering
\begin{threeparttable}
 \caption{Datasets statistics.}
\label{tab:dataset}
 \centering
\begin{tabular}{lcc}
 \toprule 
Datasets & \# Nodes & \# Edges \\
\midrule
\multicolumn{3}{c}{Node Classification} \\
\midrule
ogbn-products & 2,449,029 & 61,859,140 \\
ogbn-arxiv & 169,343 & 1,166,243 \\
ogbn-proteins & 132,524 & 39,561,252 \\
reddit & 232,965 & 114,615,892 \\
\midrule
\multicolumn{3}{c}{Link Prediction} \\
\midrule
ogbl-collab & 235,868 & 1,285,465 \\
ogbl-ppa & 576,289 & 30,326,273 \\
ogbl-ddi & 4,267 & 1,334,889 \\
\bottomrule
\end{tabular}
\end{threeparttable}
\end{table}

\begin{table*}
\centering
\begin{threeparttable}
 \caption{Baseline GNN models used in evaluation.}
\label{tab:models}
\begin{tabular}{ll}
 \toprule 
 GNN Model & Description \\
 \midrule
 \multicolumn{2}{c}{Node classification task} \\
 \midrule
GraphSage & Vanilla GraphSage with neighbor sampling. \\
GraphSage-Cluster & Vanilla GraphSage with cluster based sampling~\cite{chiang2019cluster}. \\
GAT-FLAG & GAT with FLAG~\cite{kong2020flag} enhancement. \\
GAT+BoT & GAT with bag of tricks~\cite{wang2021bag}. \\
AGDN+BoT & AGDN with bag of tricks. \\
AGDN+BoT+self-KD+C\&S & AGDN with bag of tricks, knowledge distillation and correct\&smooth\cite{huang2020combining}. \\
 \midrule
\multicolumn{2}{c}{Link prediction task} \\
 \midrule
SEAL-DGCNN & Vanilla SEAL using DGCNN~\cite{zhang2018end} as the backbone GNN.\\
GCN-full & Vanilla GCN with full graph training. \\
GraphSage-full & Vanilla Sage with full graph training. \\
GraphSage+EdgeAttr & GraphSage with edge attribute.\\
\bottomrule
\end{tabular}
\begin{tablenotes}
\scriptsize
\small
  \item GraphSage+EdgeAttr comes from https://github.com/lustoo/OGB\_link\_prediction 
\end{tablenotes}
\end{threeparttable}
\end{table*}

\begin{table*}[t]
 \centering
\begin{threeparttable}
 \caption{Performance of \method on ogbn-products, ogbn-arxiv, ogbn-proteins and reddit.}
\label{tab:node-class}
\begin{tabular}{lccccc}
 \toprule 
Dataset & & ogbn-products & ogbn-arxiv & ogbn-proteins & reddit \\
Eval Metric & &Accuracy(\%) & Accuracy(\%) & ROC-AUC(\%) & Accuracy(\%) \\
\midrule
\multirow{2}{*}{GraphSage} & Vanilla & 78.27$\pm$0.45   &71.15$\pm$1.66   &75.67$\pm$1.72   &96.19$\pm$0.08 \\
& \method & \textbf{79.88$\pm$0.34}   &  \textbf{71.77$\pm$1.18}   & \textbf{76.30$\pm$0.96}   &96.21$\pm$0.04 \\
\midrule
\multirow{2}{*}{GraphSage-Cluster} & Vanilla & 78.72$\pm$0.63 &   56.57$\pm$1.56  & 67.45$\pm$1.21   & 95.27$\pm$0.09 \\
& \method &   78.91$\pm$0.59 &   56.76$\pm$1.08 & \ \textbf{68.12$\pm$0.96}   &95.34$\pm$0.09 \\
\midrule
\multirow{2}{*}{GAT-NS} & Vanilla & 79.23$\pm$0.16   & 72.10$\pm$1.12   &81.76$\pm$0.17   & 96.12$\pm$0.02 \\
& \method &79.67$\pm$0.09   & 71.88$\pm$1.10   &81.91$\pm$0.21   & 96.45$\pm$0.05 \\
\midrule
\multirow{2}{*}{GAT-FLAG} & Vanilla &80.75$\pm$0.14   &71.56$\pm$1.11  &81.81$\pm$0.15   &95.27$\pm$0.02 \\
& \method &80.99$\pm$0.09   &71.74$\pm$1.10   &81.84$\pm$0.11   &95.68$\pm$0.03 \\

\bottomrule
\end{tabular}
\begin{tablenotes}
\scriptsize
\small
  \item Any score difference between vanilla GNN and \stmethod GNN that is greater than 0.5\% is highlighted with boldface. 
\end{tablenotes}
\end{threeparttable}
\end{table*}

\begin{table}[t]
 \centering
\begin{threeparttable}
 \caption{Performance (as measured by classification accuracy and ROC-AUC for ogbn-arxiv and ogbn-proteins, respectively) of \method combined with bag of tricks on ogbn-arxiv and ogbn-proteins.}
\label{tab:node-class-bot}
\begin{tabular}{lccc}
Dataset & Model &  & Accuracy(\%) \\
\midrule
ogbn-arxiv & \multirow{2}{*}{AGDN+BoT} & Vanilla & 74.03$\pm$0.15\\
ogbn-arxiv & & \method& 74.25$\pm$0.17 \\
\midrule
ogbn-arxiv & AGDN+BoT+ & Vanilla & 74.28$\pm$0.13 \\
ogbn-arxiv & self-KD+C\&S & \method & 74.34$\pm$0.14 \\
\midrule
ogbn-proteins &\multirow{2}{*}{GAT+BoT} & Vanilla & 87.73$\pm$0.18 \\
ogbn-proteins & & \method & 88.09$\pm$0.1 \\
 \toprule 
 \end{tabular}
\end{threeparttable}
\end{table}
\begin{table*}[t]
 \centering
\begin{threeparttable}
 \caption{Performance of \method on ogbl-collab, ogbl-ppa and ogbl-ppi. We use the hit@20, hit@50 and hit@100 as the evaluation metrics.}
\label{tab:link-prediction1}
 \centering
\begin{tabular}{lccccccc}
 \toprule 
& \multirow{2}{*}{Metric (\%)} & \multicolumn{2}{c}{ogbl-collab} & \multicolumn{2}{c}{ogbl-ppa} & \multicolumn{2}{c}{ogbl-ddi} \\
& & Vanilla GNN & \method & Vanilla GNN & \method & Vanilla GNN & \method \\
\midrule
\midrule
SEAL- & hit@20 &45.76$\pm$0.72 &46.19$\pm$0.58   &16.10$\pm$1.85 & \textbf{20.82$\pm$1.76}  & 30.75$\pm$2.12 & \textbf{31.93$\pm$3.00} \\
DGCNN& hit@50 &54.70$\pm$0.49 & 54.82$\pm$0.20  &32.58$\pm$1.42 & \textbf{37.25$\pm$0.98}  & 43.99$\pm$1.11 & 42.39$\pm$3.23 \\
& hit@100 &60.13$\pm$0.32  & 60.70$\pm$0.18  &49.36$\pm$1.24 & \textbf{56.44$\pm$0.99}  & 51.25$\pm$1.60 & 49.63$\pm$3.65 \\
\midrule
\multirow{3}{*}{GCN-full} & hit@10 & 35.94$\pm$1.60 &  36.69$\pm$0.82 & 4.00$\pm$1.46& \textbf{5.64$\pm$0.93}   & 47.82 $\pm$ 5.90 & 48.22 $\pm$ 7.00 \\
& hit@50 & 49.52$\pm$0.70 & \textbf{51.83$\pm$0.50}  & 14.23$\pm$1.81 & \textbf{18.44$\pm$1.88}  & 79.56$\pm$3.83 &  \textbf{82.56$\pm$4.03} \\
& hit@100 & 55.74$\pm$0.44 & \textbf{57.41$\pm$0.22} & 20.21$\pm$ 1.92  & \textbf{26.78$\pm$0.92}  & 87.58$\pm$1.33 &  \textbf{89.48$\pm$1.68} \\
\midrule
GraphSage- & hit@10 & 32.59$\pm$3.56  & \textbf{36.83$\pm$2.56}  & 3.68$\pm$1.02 & 3.52$\pm$1.24  & 54.27$\pm$9.86 & \textbf{60.75$\pm$4.94} \\
full& hit@50 & 51.66$\pm$0.35 & 52.62$\pm$1.04 & 15.02$\pm$1.69 & 15.55$\pm$1.92 & 82.18$\pm$4.00 & \textbf{84.58$\pm$1.89}\\
& hit@100 & 56.91$\pm$0.72 & \textbf{57.96$\pm$0.56} & 23.56$\pm$1.58 & 24.45$\pm$2.34 & 91.94$\pm$0.64 & 92.58$\pm$0.88\\
\midrule
GraphSage+ & hit@20 & -  & -  & - & -   & 87.06$\pm$4.81 & \textbf{93.28$\pm$1.61}\\
EdgeAttr & hit@50 & - & -  & -  & -  &  97.98$\pm$0.42 & 98.39$\pm$0.21\\
& hit@100 & - & - & - & -  & 98.98$\pm$0.16 & 99.21$\pm$0.08\\
\bottomrule
\end{tabular}
\begin{tablenotes}
\scriptsize
\small
  \item The evaluation metrics used in ogb learderboard are hit@50 for ogbl-collab, hit@100 for ogbl-ppa and hit@20 for ogbl-ddi.
  \item Any hit score difference between vanilla GNN and \method GNN that is greater than 1\% is highlighted with boldface. 
  \item The evaluation metrics used for ogbl-ddi when profiling GCN-full and GraphSage-full are hit@20, hit@50 and hit@100.
\end{tablenotes}
\end{threeparttable}
\end{table*}

We evaluated the effectiveness of \method by applying it to various GNN models including GCN~\cite{kipf2016semi}, Graphsage~\cite{hamilton2017inductive}, Graph Attention Network (GAT)~\cite{velivckovic2017graph}, Adaptive Graph Diffusion Networks (AGDN)~\cite{sun2020adaptive}, and SEAL~\cite{zhang2020revisiting} and their variants. Table~\ref{tab:models} presents all the baseline models. We directly followed the implementation and configuration of each baseline model from the OGB~\cite{hu2020open} leaderboard and added non-linear layer(s) into each GNN layer for \method. Table~\ref{tab:setting} presents the detail configuration of each model.
All models were trained on a single V100 GPU with 32GB memory. We report average performance over 10 runs for all models except SEAL related models. As training SEAL models is very expensive, we took 5 runs instead.

\subsection{Node classification} \label{eval-nc}
Firstly, we analyzed how \method improves the performance of GNN models on node classification tasks. Table~\ref{tab:node-class} presents the overall results.
It can be seen that \stmethod-based models outperform their baseline models in most of the cases. Notably, \method tends to performs well with GraphSage. It improves the test accuracy of GraphSage on ogbn-products and ogbn-arxiv by 1.61 and 0.62 respectively. It also improves the ROC-AUC score of GraphSage on ogbn-proteins by 0.63. 
But as the baseline performance of reddit dataset is quite high, not surprisingly, the overall improvement of \method is not significant.

We further analysis the performance of \method combined with bag of tricks~\cite{wang2021bag} on ogbn-arxiv and ogbn-proteins in Table~\ref{tab:node-class-bot}. It can be seen that \stmethod-based models outperform their vanilla counterparts. \method with AGDN+BoT+self-KD+C\&S even achieves the first place over all the methods with no extension to the input data on the ogbn-arxiv leaderboard as of the time of this submission (The forth place on the entire ogbn-arxiv leaderboard). 
\method with GAT+BoT also achieves the second place on the ogbn-proteins leaderboard with 5.83 times fewer parameters compared with the current leading method RevGNN-Wide.\footnote{\method-GAT+Bot has 11,740,552 parameters while RevGNN-Wide has 68,471,608 parameters.}

\subsection{Link prediction} \label{eval-lp}
Secondly, we analyzed how \method improves the performance of GNN models on link prediction tasks. Table~\ref{tab:link-prediction1} presents the results on the ogbl-collab, ogbl-ppa and ogbl-ppi datasets. 
As shown in the tables, the performance improvement of \method over SEAL models is significant.
\method improves the hit@20, hit@50 and hit@100 of SEAL-DGCNN by 4.72\%, 4.67\% and 7.08\% respectively on ogbl-ppa. \method with SEAL-DGCNN achieves the first place on the ogbn-ppa leaderboard with an improvement of hit@100 by 5.82\% over the current leading method  MLP+CN\&RA\&AA~\footnote{https://github.com/lustoo/OGB\_link\_prediction}.
Furthermore, \method with GraphSage+EdgeAttr achieves the first place on the ogbl-ddi leaderboard with an improvement of hit@20 by 5.47\% over the current leading method vanilla GraphSage+EdgeAttr. As GraphSage+EdgeAttr only provided the performance on ogbl-ppi, we do not compare its performance on other datasets.
\method also works with GCN and GraphSage on link prediction tasks. As shown in the tables, 
It improves the performance of GCN and GraphSage in all cases. In particular, it improves the hit@20, hit@50 and hit@100 of GCN by 1.64\%, 4.21\% and 6.57\% respectively on ogbl-ppa.


\begin{table}[t]
 \caption{Test accuracy (\%) of GraphSage and GAT with and without \method trained with different training methods on ogbn-products.}
\label{tab:node-train-method}
 \centering
\begin{tabular}{lccc}
 \toprule 
Sampling & \multirow{2}{*}{full-graph} & neighbor & cluster-based \\
Methods & & sampling & sampling\\
\midrule
GraphSage &78.27  &78.70  &78.72	 \\
GraphSage-\method &79.88  &79.11  &78.91 \\
\midrule
GAT &80.75 &79.23 &71.41 \\
GAT-\method &80.99 &79.67 &76.76\\
\bottomrule
\end{tabular}
\end{table}

\begin{table}[t]
 \caption{Test accuracy (\%) of GraphSage and GAT with different number of non-linear layers added into GNN layers on ogbn-products.}
\label{tab:node-class-ml}
 \centering
\begin{tabular}{lrrr}
 \toprule 
Model & \multicolumn{3}{c}{GraphSage}\\
Hidden-size & 128 & 256 & 512 \\
\midrule
baseline & 77.44 & 78.27 & 79.37	 \\
\method-1layer & 77.39 & 79.53 & 79.12  \\
\method-2layer & 78.79 & 79.88 & 79.94  \\
\method-4layer & 78.79 & 79.52 & 79.88  \\
\midrule
Model & \multicolumn{3}{c}{GAT} \\
Hidden-size & 64 & 128 & 256 \\
\midrule
baseline & 68.41  &79.23  &75.26	 \\
\method-1layer & 69.72  &79.67  &77.53	 \\
\method-2layer & 69.86  &78.26  &78.76	 \\
\method-4layer & 69.41  &78.23  &78.61	 \\
\bottomrule
\end{tabular}
\end{table}

\begin{table}
 \caption{The parameter size of each model in Table~\ref{tab:node-class-ml}.}
\label{tab:node-class-ml-size}
 \centering
\begin{tabular}{lrrr}
 \toprule 
Model & \multicolumn{3}{c}{GraphSage}\\
Hidden-size & 128 & 256 & 512 \\
\midrule
baseline & 70,703 & 206,895 & 675,887	 \\
\method-1layer & 87,215 & 272,687 & 938,543  \\
\method-2layer & 103,727 & 338,479 &  1,201,199 \\
\method-4layer & 136,751 & 470,063 & 1,726,511  \\
\midrule
Model & \multicolumn{3}{c}{GAT} \\
Hidden-size & 64 & 128 & 256 \\
\midrule
baseline & 510,056  & 1,543,272  & 5,182,568	 \\
\method-1layer & 514,152  & 1,559,656  & 5,248,104	 \\
\method-2layer & 518,248  & 1,576,040  & 5,313,640 \\
\method-4layer & 526,440  & 1,608,808 & 5,444,712 \\
\bottomrule
\end{tabular}
\end{table}

\begin{table}
 \caption{The single training epoch time of GraphSage with different number of non-linear layers added into GNN layers on ogbn-products.}
\label{tab:node-class-time}
 \centering
\begin{tabular}{lrrr}
 \toprule 
Model & \multicolumn{3}{c}{GraphSage (secs)}\\
Hidden-size & 128 & 256 & 512 \\
\midrule
baseline & 18.71$\pm$0.41 & 20.71$\pm$0.74 & 25.40$\pm$0.35	 \\
\method-1layer & 19.57$\pm$1.04 & 20.98$\pm$0.47 & 29.07$\pm$0.69  \\
\method-2layer & 19.25$\pm$0.87 & 21.36$\pm$0.48 & 30.01$\pm$0.13 \\
\method-4layer & 19.79$\pm$0.72 & 24.41$\pm$0.38 & 32.33$\pm$0.19  \\
\bottomrule
\end{tabular}
\end{table}

\subsection{\method with Different Training Methods} \label{eval-train}
Finally, we presents the effectiveness of using \method with different training methods including full-graph training, neighbor sampling and cluster-based sampling. Table~\ref{tab:node-train-method} presents the results. It can be seen that \method improves the performance of GraphSage and GAT with all kinds of training methods on ogbn-products. 
It is worth mentioning that \method also works with local subgraph sampling method proposed by SEAL~\cite{zhang2018link} as shown in Section~\ref{eval-lp}.


\subsection{Effectiveness of Multiple \method Layers}
\label{eval-mtl}
We studied the effectiveness of adding multiple non-linear layers to GNN layers on ogbn-products using GraphSage and GAT. Table~\ref{tab:node-class-ml} presents the results. The baseline model is a three-layer GNN model. We applied 1, 2 or 4 non-linear layers to each hidden GNN layer denoted as \stmethod-1layer, \stmethod-2layer and \stmethod-4layer respectively. The GAT models use eight attention heads and all heads share the same NGNN layer(s).
Table~\ref{tab:node-class-ml} presents the result. 
As shown in the table, \stmethod-2layer always performed best with different hidden sizes in most of the cases. This reveals that adding non-linear layers can be effective, but the effect may vanish significantly when we continuously add more layers. The reason is straightforward, given that adding more non-linear layers can eventually cause overfitting. 

We also observe that deeper models can achieve better performance with many fewer trainable parameters than wider models. Table~\ref{tab:node-class-ml-size} presents the model parameter size of each model.
As shown in the table, the parameter size of GraphSage with \stmethod-2layer and a hidden size of 256 is 338,479 which is 2$\times$ smaller than the parameter size of vanilla GraphSage with a hidden-size of 512, i.e., 675,887. And its performance is much better than vanilla GraphSage with a hidden size of 512. 

Furthermore, we also observe that adding \stmethod layers only slightly increase the model training time. Table~\ref{tab:node-class-time} presents the single training epoch time of GraphSage under different configurations. As shown in the table, the epoch time of GraphSage with \stmethod-2layer and a hidden size of 256 is only 3.1\% longer than that of vanilla GraphSage with the same hidden size. However the corresponding parameter size is 1.63$\times$ larger.  


\begin{table}
 \caption{Test accuracy (\%) of GraphSage and GAT when applying \method on different GNN layers on ogbn-product.}
\label{tab:input-hidden}
 \centering
 \begin{tabular}{lrr}
 \toprule 
 & GraphSage & GAT \\
 \midrule
baseline  & 78.27 & 79.23 \\
\method-all & 79.88 & 79.49 \\
\method-input   & 79.81 & 78.87 \\
\method-hidden   & 79.91 & 79.68 \\
\method-output   & 78.60 & 78.45\\
\bottomrule
\end{tabular}
\end{table}

\subsection{Effectiveness of Applying \method to Different GNN Layers} \label{eval-dgnn}
Finally, we studied the effectiveness of applying \method to only the input GNN layer (\stmethod-input), only the hidden GNN layers (\stmethod-hidden), only the output GNN layer (\stmethod-output) and all the GNN layers on the ogbn-products dataset using GraphSage and GAT. 
The baseline model is a three-layer GNN model. The hidden dimension size is 256 and 128 for GraphSage and GAT respectively.
Table~\ref{tab:input-hidden} presents the results. As the table shows, only applying \method to the output GNN layer brings little or no benefit.
While applying \method to hidden and input GNN layers can improve the model performance, especially applying \method to hidden layers.
It demonstrates that the benefit of \method mainly comes from adding additional non-linear layers into the input and hidden GNN layers.





\begin{table*}[t]
 \centering
\begin{threeparttable}
 \caption{Model settings of \method models. The column of \method position presents where we put the non-linear layers. hidden-only means only applying \method to the hidden GNN layers, input-only means only applying \method to the input layer, all-layer means applying \method to all the GNN layers.. The column of \method setting presents how we organize each \method layer. For example, 1-relu+1-sigmoid means \method contains one feedforward neural network with ReLU as its activation function followed by another feedforward neural network with Sigmoid as its activation function and 2-relu means \method contains two feedforward neural network layers with ReLU as the activation function of each layer.}
\label{tab:setting}
 \centering
\begin{tabular}{llccccl}
 \toprule 
Dataset & Model & hidden size & layers & aggregation & \method position & \method setting \\
\midrule
\multicolumn{6}{c}{Node classification tasks} \\
\midrule
ogbn-product & GraphSage & 256 & 3 & mean & hidden-only & 1-relu+1-sigmoid \\
ogbn-product & GraphSage-cluster & 256 & 3 & mean & hidden-only & 1-relu+1-sigmoid\\
ogbn-product & GAT-flag & 256 & 3 & sum & hidden-only & 1-relu+1-sigmoid\\
ogbn-product & GAT-ns & 256 & 3 & sum & hidden-only & 1-relu+1-sigmoid\\
ogbn-arxiv & GraphSage & 256 & 3 & mean & hidden-only & 1-relu+1-sigmoid\\
ogbn-arxiv & GraphSage-cluster & 256 & 3 & mean & hidden-only & 1-relu+1-sigmoid\\
ogbn-arxiv & GAT-flag & 256 & 3 & sum & hidden-only & 1-relu+1-sigmoid\\
ogbn-arxiv & GAT+BoT & 120 & 6 & sum & hidden-only & 2-relu \\
ogbn-arxiv & AGDN+BoT & 256 & 3 & GAT-HA & hidden-only & 1-relu \\
ogbn-arxiv & AGDN+BoT+self-KD+C\&S & 256 & 3 & GAT-HA & hidden-only & 1-relu \\
ogbn-protein & GraphSage & 256 & 3 & mean & hidden-only & 1-relu\\
ogbn-protein & GraphSage-cluster & 256 & 3 & mean & hidden-only & 1-relu\\
ogbn-protein & GAT-flag & 256 & 3 & sum & hidden-only & 1-relu\\
ogbn-protein & GAT-ns & 256 & 3 & sum & hidden-only & 1-relu\\
reddit & GraphSage & 256 & 3 & mean & hidden-only & 1-relu+1-sigmoid\\
reddit & GraphSage-cluster & 256 & 3 & mean & hidden-only & 1-relu+1-sigmoid\\
reddit & GAT-flag & 256 & 3 & sum & hidden-only & 1-relu+1-sigmoid\\
reddit & GAT-ns & 256 & 3 & sum & hidden-only & 1-relu+1-sigmoid\\
\midrule
\multicolumn{6}{c}{Link prediction tasks} \\
\midrule
ogbl-collab & Seal-DGCNN & 256 & 3 & sum & all-layers & 1-relu\\
ogbl-collab & GCN-full & 256 & 3 & mean & hidden-only & 2-relu\\
ogbl-collab & GraphSage-full & 256 & 3 & mean & hidden-only & 2-relu\\
ogbl-ppa & Seal-DGCNN & 32 & 3 & sum & all-layers & 1-relu\\
ogbl-ppa & GCN-full & 256 & 3 & mean & all-layers & 2-relu\\
ogbl-ppa & GraphSage-full & 256 & 3 & mean & all-layers & 2-relu\\
ogbl-ppi & Seal-DGCNN & 32 & 3 & sum & hidden-layers & 1-relu\\
ogbl-ppi & GCN-full & 256 & 2 &  mean & input-only & 1-relu \\
ogbl-ppi & GraphSage-full & 256 & 2 &  mean & input-only & 1-relu \\
ogbl-ppi & GraphSage+EdgeAttr & 512 & 2 & mean & all-layers & 2-relu \\
\bottomrule
\end{tabular}
\end{threeparttable}
\end{table*}

\section{Conclusion and Future Work}
We present \stmethod, a model agnostic methodology that allows arbitrary GNN models to increase their model capacity by inserting non-linear feedforward neural network layer(s) inside GNN layers. Moreover, unlike existing deep GNN approaches, \method does not have large memory overhead and can work with various training methods including neighbor sampling, graph clustering and  local subgraph sampling. 
Empirically, we demonstrate that \method can work with various GNN models on both node classification and link prediction tasks and achieve state-of-the-art results. Future work includes evaluating \method on more GNN models and investigating whether \method can work on broader graph-related prediction tasks. We also plan to explore methodologies to make a single GNN layer deeper in the future.

\bibliographystyle{ACM-Reference-Format}
\bibliography{sample-sigconf.bib}


\begin{thebibliography}{43}


\ifx \showCODEN    \undefined \def \showCODEN     #1{\unskip}     \fi
\ifx \showDOI      \undefined \def \showDOI       #1{#1}\fi
\ifx \showISBNx    \undefined \def \showISBNx     #1{\unskip}     \fi
\ifx \showISBNxiii \undefined \def \showISBNxiii  #1{\unskip}     \fi
\ifx \showISSN     \undefined \def \showISSN      #1{\unskip}     \fi
\ifx \showLCCN     \undefined \def \showLCCN      #1{\unskip}     \fi
\ifx \shownote     \undefined \def \shownote      #1{#1}          \fi
\ifx \showarticletitle \undefined \def \showarticletitle #1{#1}   \fi
\ifx \showURL      \undefined \def \showURL       {\relax}        \fi
\providecommand\bibfield[2]{#2}
\providecommand\bibinfo[2]{#2}
\providecommand\natexlab[1]{#1}
\providecommand\showeprint[2][]{arXiv:#2}

\bibitem[\protect\citeauthoryear{Berg, Kipf, and Welling}{Berg
  et~al\mbox{.}}{2017}]%
        {berg2017graph}
\bibfield{author}{\bibinfo{person}{Rianne van~den Berg},
  \bibinfo{person}{Thomas~N Kipf}, {and} \bibinfo{person}{Max Welling}.}
  \bibinfo{year}{2017}\natexlab{}.
\newblock \showarticletitle{Graph convolutional matrix completion}.
\newblock \bibinfo{journal}{\emph{arXiv preprint arXiv:1706.02263}}
  (\bibinfo{year}{2017}).
\newblock


\bibitem[\protect\citeauthoryear{Brown, Mann, Ryder, Subbiah, Kaplan, Dhariwal,
  Neelakantan, Shyam, Sastry, Askell, et~al\mbox{.}}{Brown
  et~al\mbox{.}}{2020}]%
        {brown2020language}
\bibfield{author}{\bibinfo{person}{Tom~B Brown}, \bibinfo{person}{Benjamin
  Mann}, \bibinfo{person}{Nick Ryder}, \bibinfo{person}{Melanie Subbiah},
  \bibinfo{person}{Jared Kaplan}, \bibinfo{person}{Prafulla Dhariwal},
  \bibinfo{person}{Arvind Neelakantan}, \bibinfo{person}{Pranav Shyam},
  \bibinfo{person}{Girish Sastry}, \bibinfo{person}{Amanda Askell},
  {et~al\mbox{.}}} \bibinfo{year}{2020}\natexlab{}.
\newblock \showarticletitle{Language models are few-shot learners}.
\newblock \bibinfo{journal}{\emph{arXiv preprint arXiv:2005.14165}}
  (\bibinfo{year}{2020}).
\newblock


\bibitem[\protect\citeauthoryear{Chen, Lin, Li, Li, Zhou, and Sun}{Chen
  et~al\mbox{.}}{2020}]%
        {chen2020measuring}
\bibfield{author}{\bibinfo{person}{Deli Chen}, \bibinfo{person}{Yankai Lin},
  \bibinfo{person}{Wei Li}, \bibinfo{person}{Peng Li}, \bibinfo{person}{Jie
  Zhou}, {and} \bibinfo{person}{Xu Sun}.} \bibinfo{year}{2020}\natexlab{}.
\newblock \showarticletitle{Measuring and relieving the over-smoothing problem
  for graph neural networks from the topological view}. In
  \bibinfo{booktitle}{\emph{Proceedings of the AAAI Conference on Artificial
  Intelligence}}, Vol.~\bibinfo{volume}{34}. \bibinfo{pages}{3438--3445}.
\newblock


\bibitem[\protect\citeauthoryear{Chen, Ma, and Xiao}{Chen
  et~al\mbox{.}}{2018}]%
        {chen2018fastgcn}
\bibfield{author}{\bibinfo{person}{Jie Chen}, \bibinfo{person}{Tengfei Ma},
  {and} \bibinfo{person}{Cao Xiao}.} \bibinfo{year}{2018}\natexlab{}.
\newblock \showarticletitle{Fastgcn: fast learning with graph convolutional
  networks via importance sampling}.
\newblock \bibinfo{journal}{\emph{arXiv preprint arXiv:1801.10247}}
  (\bibinfo{year}{2018}).
\newblock


\bibitem[\protect\citeauthoryear{Chiang, Liu, Si, Li, Bengio, and Hsieh}{Chiang
  et~al\mbox{.}}{2019}]%
        {chiang2019cluster}
\bibfield{author}{\bibinfo{person}{Wei-Lin Chiang}, \bibinfo{person}{Xuanqing
  Liu}, \bibinfo{person}{Si Si}, \bibinfo{person}{Yang Li},
  \bibinfo{person}{Samy Bengio}, {and} \bibinfo{person}{Cho-Jui Hsieh}.}
  \bibinfo{year}{2019}\natexlab{}.
\newblock \showarticletitle{Cluster-gcn: An efficient algorithm for training
  deep and large graph convolutional networks}. In
  \bibinfo{booktitle}{\emph{Proceedings of the 25th ACM SIGKDD International
  Conference on Knowledge Discovery \& Data Mining}}.
  \bibinfo{pages}{257--266}.
\newblock


\bibitem[\protect\citeauthoryear{Fan, Ma, Li, He, Zhao, Tang, and Yin}{Fan
  et~al\mbox{.}}{2019}]%
        {fan2019graph}
\bibfield{author}{\bibinfo{person}{Wenqi Fan}, \bibinfo{person}{Yao Ma},
  \bibinfo{person}{Qing Li}, \bibinfo{person}{Yuan He}, \bibinfo{person}{Eric
  Zhao}, \bibinfo{person}{Jiliang Tang}, {and} \bibinfo{person}{Dawei Yin}.}
  \bibinfo{year}{2019}\natexlab{}.
\newblock \showarticletitle{Graph neural networks for social recommendation}.
  In \bibinfo{booktitle}{\emph{The World Wide Web Conference}}.
  \bibinfo{pages}{417--426}.
\newblock


\bibitem[\protect\citeauthoryear{Gilmer, Schoenholz, Riley, Vinyals, and
  Dahl}{Gilmer et~al\mbox{.}}{2017}]%
        {gilmer2017neural}
\bibfield{author}{\bibinfo{person}{Justin Gilmer}, \bibinfo{person}{Samuel~S
  Schoenholz}, \bibinfo{person}{Patrick~F Riley}, \bibinfo{person}{Oriol
  Vinyals}, {and} \bibinfo{person}{George~E Dahl}.}
  \bibinfo{year}{2017}\natexlab{}.
\newblock \showarticletitle{Neural message passing for quantum chemistry}. In
  \bibinfo{booktitle}{\emph{International conference on machine learning}}.
  PMLR, \bibinfo{pages}{1263--1272}.
\newblock


\bibitem[\protect\citeauthoryear{Hamilton, Ying, and Leskovec}{Hamilton
  et~al\mbox{.}}{2017}]%
        {hamilton2017inductive}
\bibfield{author}{\bibinfo{person}{William~L Hamilton}, \bibinfo{person}{Rex
  Ying}, {and} \bibinfo{person}{Jure Leskovec}.}
  \bibinfo{year}{2017}\natexlab{}.
\newblock \showarticletitle{Inductive representation learning on large graphs}.
  In \bibinfo{booktitle}{\emph{Proceedings of the 31st International Conference
  on Neural Information Processing Systems}}. \bibinfo{pages}{1025--1035}.
\newblock


\bibitem[\protect\citeauthoryear{He, Zhang, Ren, and Sun}{He
  et~al\mbox{.}}{2016}]%
        {he2016deep}
\bibfield{author}{\bibinfo{person}{Kaiming He}, \bibinfo{person}{Xiangyu
  Zhang}, \bibinfo{person}{Shaoqing Ren}, {and} \bibinfo{person}{Jian Sun}.}
  \bibinfo{year}{2016}\natexlab{}.
\newblock \showarticletitle{Deep residual learning for image recognition}. In
  \bibinfo{booktitle}{\emph{Proceedings of the IEEE conference on computer
  vision and pattern recognition}}. \bibinfo{pages}{770--778}.
\newblock


\bibitem[\protect\citeauthoryear{Hu, Fey, Zitnik, Dong, Ren, Liu, Catasta, and
  Leskovec}{Hu et~al\mbox{.}}{2020}]%
        {hu2020open}
\bibfield{author}{\bibinfo{person}{Weihua Hu}, \bibinfo{person}{Matthias Fey},
  \bibinfo{person}{Marinka Zitnik}, \bibinfo{person}{Yuxiao Dong},
  \bibinfo{person}{Hongyu Ren}, \bibinfo{person}{Bowen Liu},
  \bibinfo{person}{Michele Catasta}, {and} \bibinfo{person}{Jure Leskovec}.}
  \bibinfo{year}{2020}\natexlab{}.
\newblock \showarticletitle{Open graph benchmark: Datasets for machine learning
  on graphs}.
\newblock \bibinfo{journal}{\emph{arXiv preprint arXiv:2005.00687}}
  (\bibinfo{year}{2020}).
\newblock


\bibitem[\protect\citeauthoryear{Huang and Carley}{Huang and Carley}{2019}]%
        {huang2019residual}
\bibfield{author}{\bibinfo{person}{Binxuan Huang} {and}
  \bibinfo{person}{Kathleen~M Carley}.} \bibinfo{year}{2019}\natexlab{}.
\newblock \showarticletitle{Residual or gate? towards deeper graph neural
  networks for inductive graph representation learning}.
\newblock \bibinfo{journal}{\emph{arXiv preprint arXiv:1904.08035}}
  (\bibinfo{year}{2019}).
\newblock


\bibitem[\protect\citeauthoryear{Huang, Liu, Van Der~Maaten, and
  Weinberger}{Huang et~al\mbox{.}}{2017}]%
        {huang2017cvpr}
\bibfield{author}{\bibinfo{person}{Gao Huang}, \bibinfo{person}{Zhuang Liu},
  \bibinfo{person}{Laurens Van Der~Maaten}, {and} \bibinfo{person}{Kilian~Q.
  Weinberger}.} \bibinfo{year}{2017}\natexlab{}.
\newblock \showarticletitle{Densely Connected Convolutional Networks}. In
  \bibinfo{booktitle}{\emph{2017 IEEE Conference on Computer Vision and Pattern
  Recognition (CVPR)}}. \bibinfo{pages}{2261--2269}.
\newblock
\urldef\tempurl%
\url{https://doi.org/10.1109/CVPR.2017.243}
\showDOI{\tempurl}


\bibitem[\protect\citeauthoryear{Huang, He, Singh, Lim, and Benson}{Huang
  et~al\mbox{.}}{2020}]%
        {huang2020combining}
\bibfield{author}{\bibinfo{person}{Qian Huang}, \bibinfo{person}{Horace He},
  \bibinfo{person}{Abhay Singh}, \bibinfo{person}{Ser-Nam Lim}, {and}
  \bibinfo{person}{Austin~R Benson}.} \bibinfo{year}{2020}\natexlab{}.
\newblock \showarticletitle{Combining label propagation and simple models
  out-performs graph neural networks}.
\newblock \bibinfo{journal}{\emph{arXiv preprint arXiv:2010.13993}}
  (\bibinfo{year}{2020}).
\newblock


\bibitem[\protect\citeauthoryear{Kipf and Welling}{Kipf and Welling}{2016}]%
        {kipf2016semi}
\bibfield{author}{\bibinfo{person}{Thomas~N Kipf} {and} \bibinfo{person}{Max
  Welling}.} \bibinfo{year}{2016}\natexlab{}.
\newblock \showarticletitle{Semi-supervised classification with graph
  convolutional networks}.
\newblock \bibinfo{journal}{\emph{arXiv preprint arXiv:1609.02907}}
  (\bibinfo{year}{2016}).
\newblock


\bibitem[\protect\citeauthoryear{Kong, Li, Ding, Wu, Zhu, Ghanem, Taylor, and
  Goldstein}{Kong et~al\mbox{.}}{2020}]%
        {kong2020flag}
\bibfield{author}{\bibinfo{person}{Kezhi Kong}, \bibinfo{person}{Guohao Li},
  \bibinfo{person}{Mucong Ding}, \bibinfo{person}{Zuxuan Wu},
  \bibinfo{person}{Chen Zhu}, \bibinfo{person}{Bernard Ghanem},
  \bibinfo{person}{Gavin Taylor}, {and} \bibinfo{person}{Tom Goldstein}.}
  \bibinfo{year}{2020}\natexlab{}.
\newblock \showarticletitle{Flag: Adversarial data augmentation for graph
  neural networks}.
\newblock \bibinfo{journal}{\emph{arXiv preprint arXiv:2010.09891}}
  (\bibinfo{year}{2020}).
\newblock


\bibitem[\protect\citeauthoryear{Li, Qin, Liu, Yang, and Li}{Li
  et~al\mbox{.}}{2019b}]%
        {li2019spam}
\bibfield{author}{\bibinfo{person}{Ao Li}, \bibinfo{person}{Zhou Qin},
  \bibinfo{person}{Runshi Liu}, \bibinfo{person}{Yiqun Yang}, {and}
  \bibinfo{person}{Dong Li}.} \bibinfo{year}{2019}\natexlab{b}.
\newblock \showarticletitle{Spam review detection with graph convolutional
  networks}. In \bibinfo{booktitle}{\emph{Proceedings of the 28th ACM
  International Conference on Information and Knowledge Management}}.
  \bibinfo{pages}{2703--2711}.
\newblock


\bibitem[\protect\citeauthoryear{Li, M{\"u}ller, Ghanem, and Koltun}{Li
  et~al\mbox{.}}{2021}]%
        {li2021training}
\bibfield{author}{\bibinfo{person}{Guohao Li}, \bibinfo{person}{Matthias
  M{\"u}ller}, \bibinfo{person}{Bernard Ghanem}, {and} \bibinfo{person}{Vladlen
  Koltun}.} \bibinfo{year}{2021}\natexlab{}.
\newblock \showarticletitle{Training Graph Neural Networks with 1000 Layers}.
\newblock \bibinfo{journal}{\emph{arXiv preprint arXiv:2106.07476}}
  (\bibinfo{year}{2021}).
\newblock


\bibitem[\protect\citeauthoryear{Li, Muller, Thabet, and Ghanem}{Li
  et~al\mbox{.}}{2019a}]%
        {li2019deepgcns}
\bibfield{author}{\bibinfo{person}{Guohao Li}, \bibinfo{person}{Matthias
  Muller}, \bibinfo{person}{Ali Thabet}, {and} \bibinfo{person}{Bernard
  Ghanem}.} \bibinfo{year}{2019}\natexlab{a}.
\newblock \showarticletitle{Deepgcns: Can gcns go as deep as cnns?}. In
  \bibinfo{booktitle}{\emph{Proceedings of the IEEE/CVF International
  Conference on Computer Vision}}. \bibinfo{pages}{9267--9276}.
\newblock


\bibitem[\protect\citeauthoryear{Li, Xiong, Thabet, and Ghanem}{Li
  et~al\mbox{.}}{2020}]%
        {li2020deepergcn}
\bibfield{author}{\bibinfo{person}{Guohao Li}, \bibinfo{person}{Chenxin Xiong},
  \bibinfo{person}{Ali Thabet}, {and} \bibinfo{person}{Bernard Ghanem}.}
  \bibinfo{year}{2020}\natexlab{}.
\newblock \showarticletitle{Deepergcn: All you need to train deeper gcns}.
\newblock \bibinfo{journal}{\emph{arXiv preprint arXiv:2006.07739}}
  (\bibinfo{year}{2020}).
\newblock


\bibitem[\protect\citeauthoryear{Lin, Chen, and Yan}{Lin et~al\mbox{.}}{2013}]%
        {lin2013network}
\bibfield{author}{\bibinfo{person}{Min Lin}, \bibinfo{person}{Qiang Chen},
  {and} \bibinfo{person}{Shuicheng Yan}.} \bibinfo{year}{2013}\natexlab{}.
\newblock \showarticletitle{Network in network}.
\newblock \bibinfo{journal}{\emph{arXiv preprint arXiv:1312.4400}}
  (\bibinfo{year}{2013}).
\newblock


\bibitem[\protect\citeauthoryear{Liu, Dou, Yu, Deng, and Peng}{Liu
  et~al\mbox{.}}{2020}]%
        {liu2020alleviating}
\bibfield{author}{\bibinfo{person}{Zhiwei Liu}, \bibinfo{person}{Yingtong Dou},
  \bibinfo{person}{Philip~S Yu}, \bibinfo{person}{Yutong Deng}, {and}
  \bibinfo{person}{Hao Peng}.} \bibinfo{year}{2020}\natexlab{}.
\newblock \showarticletitle{Alleviating the inconsistency problem of applying
  graph neural network to fraud detection}. In
  \bibinfo{booktitle}{\emph{Proceedings of the 43rd International ACM SIGIR
  Conference on Research and Development in Information Retrieval}}.
  \bibinfo{pages}{1569--1572}.
\newblock


\bibitem[\protect\citeauthoryear{Nt and Maehara}{Nt and Maehara}{2019}]%
        {nt2019revisiting}
\bibfield{author}{\bibinfo{person}{Hoang Nt} {and} \bibinfo{person}{Takanori
  Maehara}.} \bibinfo{year}{2019}\natexlab{}.
\newblock \showarticletitle{Revisiting graph neural networks: All we have is
  low-pass filters}.
\newblock \bibinfo{journal}{\emph{arXiv preprint arXiv:1905.09550}}
  (\bibinfo{year}{2019}).
\newblock


\bibitem[\protect\citeauthoryear{Oono and Suzuki}{Oono and Suzuki}{2019}]%
        {oono2019graph}
\bibfield{author}{\bibinfo{person}{Kenta Oono} {and} \bibinfo{person}{Taiji
  Suzuki}.} \bibinfo{year}{2019}\natexlab{}.
\newblock \showarticletitle{Graph neural networks exponentially lose expressive
  power for node classification}.
\newblock \bibinfo{journal}{\emph{arXiv preprint arXiv:1905.10947}}
  (\bibinfo{year}{2019}).
\newblock


\bibitem[\protect\citeauthoryear{Rossi, Chamberlain, Frasca, Eynard, Monti, and
  Bronstein}{Rossi et~al\mbox{.}}{2020}]%
        {rossi2020temporal}
\bibfield{author}{\bibinfo{person}{Emanuele Rossi}, \bibinfo{person}{Ben
  Chamberlain}, \bibinfo{person}{Fabrizio Frasca}, \bibinfo{person}{Davide
  Eynard}, \bibinfo{person}{Federico Monti}, {and} \bibinfo{person}{Michael
  Bronstein}.} \bibinfo{year}{2020}\natexlab{}.
\newblock \showarticletitle{Temporal graph networks for deep learning on
  dynamic graphs}.
\newblock \bibinfo{journal}{\emph{arXiv preprint arXiv:2006.10637}}
  (\bibinfo{year}{2020}).
\newblock


\bibitem[\protect\citeauthoryear{Scarselli, Gori, Tsoi, Hagenbuchner, and
  Monfardini}{Scarselli et~al\mbox{.}}{2008}]%
        {scarselli2008graph}
\bibfield{author}{\bibinfo{person}{Franco Scarselli}, \bibinfo{person}{Marco
  Gori}, \bibinfo{person}{Ah~Chung Tsoi}, \bibinfo{person}{Markus
  Hagenbuchner}, {and} \bibinfo{person}{Gabriele Monfardini}.}
  \bibinfo{year}{2008}\natexlab{}.
\newblock \showarticletitle{The graph neural network model}.
\newblock \bibinfo{journal}{\emph{IEEE transactions on neural networks}}
  \bibinfo{volume}{20}, \bibinfo{number}{1} (\bibinfo{year}{2008}),
  \bibinfo{pages}{61--80}.
\newblock


\bibitem[\protect\citeauthoryear{Schlichtkrull, Kipf, Bloem, Van Den~Berg,
  Titov, and Welling}{Schlichtkrull et~al\mbox{.}}{2018}]%
        {schlichtkrull2018modeling}
\bibfield{author}{\bibinfo{person}{Michael Schlichtkrull},
  \bibinfo{person}{Thomas~N Kipf}, \bibinfo{person}{Peter Bloem},
  \bibinfo{person}{Rianne Van Den~Berg}, \bibinfo{person}{Ivan Titov}, {and}
  \bibinfo{person}{Max Welling}.} \bibinfo{year}{2018}\natexlab{}.
\newblock \showarticletitle{Modeling relational data with graph convolutional
  networks}. In \bibinfo{booktitle}{\emph{European semantic web conference}}.
  Springer, \bibinfo{pages}{593--607}.
\newblock


\bibitem[\protect\citeauthoryear{Simonyan and Zisserman}{Simonyan and
  Zisserman}{2014}]%
        {simonyan2014very}
\bibfield{author}{\bibinfo{person}{Karen Simonyan} {and}
  \bibinfo{person}{Andrew Zisserman}.} \bibinfo{year}{2014}\natexlab{}.
\newblock \showarticletitle{Very deep convolutional networks for large-scale
  image recognition}.
\newblock \bibinfo{journal}{\emph{arXiv preprint arXiv:1409.1556}}
  (\bibinfo{year}{2014}).
\newblock


\bibitem[\protect\citeauthoryear{Sun and Wu}{Sun and Wu}{2020}]%
        {sun2020adaptive}
\bibfield{author}{\bibinfo{person}{Chuxiong Sun} {and} \bibinfo{person}{Guoshi
  Wu}.} \bibinfo{year}{2020}\natexlab{}.
\newblock \showarticletitle{Adaptive graph diffusion networks with hop-wise
  attention}.
\newblock \bibinfo{journal}{\emph{arXiv preprint arXiv:2012.15024}}
  (\bibinfo{year}{2020}).
\newblock


\bibitem[\protect\citeauthoryear{Veli{\v{c}}kovi{\'c}, Cucurull, Casanova,
  Romero, Lio, and Bengio}{Veli{\v{c}}kovi{\'c} et~al\mbox{.}}{2017}]%
        {velivckovic2017graph}
\bibfield{author}{\bibinfo{person}{Petar Veli{\v{c}}kovi{\'c}},
  \bibinfo{person}{Guillem Cucurull}, \bibinfo{person}{Arantxa Casanova},
  \bibinfo{person}{Adriana Romero}, \bibinfo{person}{Pietro Lio}, {and}
  \bibinfo{person}{Yoshua Bengio}.} \bibinfo{year}{2017}\natexlab{}.
\newblock \showarticletitle{Graph attention networks}.
\newblock \bibinfo{journal}{\emph{arXiv preprint arXiv:1710.10903}}
  (\bibinfo{year}{2017}).
\newblock


\bibitem[\protect\citeauthoryear{Wang, Wen, Wu, Huang, and Xion}{Wang
  et~al\mbox{.}}{2019}]%
        {wang2019fdgars}
\bibfield{author}{\bibinfo{person}{Jianyu Wang}, \bibinfo{person}{Rui Wen},
  \bibinfo{person}{Chunming Wu}, \bibinfo{person}{Yu Huang}, {and}
  \bibinfo{person}{Jian Xion}.} \bibinfo{year}{2019}\natexlab{}.
\newblock \showarticletitle{Fdgars: Fraudster detection via graph convolutional
  networks in online app review system}. In \bibinfo{booktitle}{\emph{Companion
  Proceedings of The 2019 World Wide Web Conference}}.
  \bibinfo{pages}{310--316}.
\newblock


\bibitem[\protect\citeauthoryear{Wang, Zheng, Ye, Gan, Li, Song, Zhou, Ma, Yu,
  Gai, Xiao, He, Karypis, Li, and Zhang}{Wang et~al\mbox{.}}{2020}]%
        {wang2020deep}
\bibfield{author}{\bibinfo{person}{Minjie Wang}, \bibinfo{person}{Da Zheng},
  \bibinfo{person}{Zihao Ye}, \bibinfo{person}{Quan Gan},
  \bibinfo{person}{Mufei Li}, \bibinfo{person}{Xiang Song},
  \bibinfo{person}{Jinjing Zhou}, \bibinfo{person}{Chao Ma},
  \bibinfo{person}{Lingfan Yu}, \bibinfo{person}{Yu Gai},
  \bibinfo{person}{Tianjun Xiao}, \bibinfo{person}{Tong He},
  \bibinfo{person}{George Karypis}, \bibinfo{person}{Jinyang Li}, {and}
  \bibinfo{person}{Zheng Zhang}.} \bibinfo{year}{2020}\natexlab{}.
\newblock \bibinfo{title}{Deep Graph Library: A Graph-Centric,
  Highly-Performant Package for Graph Neural Networks}.
\newblock
\newblock
\showeprint[arxiv]{1909.01315}~[cs.LG]


\bibitem[\protect\citeauthoryear{Wang, Jin, Zhang, Yu, Zhang, and Wipf}{Wang
  et~al\mbox{.}}{2021}]%
        {wang2021bag}
\bibfield{author}{\bibinfo{person}{Yangkun Wang}, \bibinfo{person}{Jiarui Jin},
  \bibinfo{person}{Weinan Zhang}, \bibinfo{person}{Yong Yu},
  \bibinfo{person}{Zheng Zhang}, {and} \bibinfo{person}{David Wipf}.}
  \bibinfo{year}{2021}\natexlab{}.
\newblock \showarticletitle{Bag of Tricks for Node Classification with Graph
  Neural Networks}.
\newblock \bibinfo{journal}{\emph{arXiv preprint arXiv:2103.13355}}
  (\bibinfo{year}{2021}).
\newblock


\bibitem[\protect\citeauthoryear{Wu, Souza, Zhang, Fifty, Yu, and
  Weinberger}{Wu et~al\mbox{.}}{2019}]%
        {wu2019simplifying}
\bibfield{author}{\bibinfo{person}{Felix Wu}, \bibinfo{person}{Amauri Souza},
  \bibinfo{person}{Tianyi Zhang}, \bibinfo{person}{Christopher Fifty},
  \bibinfo{person}{Tao Yu}, {and} \bibinfo{person}{Kilian Weinberger}.}
  \bibinfo{year}{2019}\natexlab{}.
\newblock \showarticletitle{Simplifying graph convolutional networks}. In
  \bibinfo{booktitle}{\emph{International conference on machine learning}}.
  PMLR, \bibinfo{pages}{6861--6871}.
\newblock


\bibitem[\protect\citeauthoryear{Xu, Hu, Leskovec, and Jegelka}{Xu
  et~al\mbox{.}}{2018}]%
        {xu2018powerful}
\bibfield{author}{\bibinfo{person}{Keyulu Xu}, \bibinfo{person}{Weihua Hu},
  \bibinfo{person}{Jure Leskovec}, {and} \bibinfo{person}{Stefanie Jegelka}.}
  \bibinfo{year}{2018}\natexlab{}.
\newblock \showarticletitle{How powerful are graph neural networks?}
\newblock \bibinfo{journal}{\emph{arXiv preprint arXiv:1810.00826}}
  (\bibinfo{year}{2018}).
\newblock


\bibitem[\protect\citeauthoryear{Ying, He, Chen, Eksombatchai, Hamilton, and
  Leskovec}{Ying et~al\mbox{.}}{2018}]%
        {ying2018graph}
\bibfield{author}{\bibinfo{person}{Rex Ying}, \bibinfo{person}{Ruining He},
  \bibinfo{person}{Kaifeng Chen}, \bibinfo{person}{Pong Eksombatchai},
  \bibinfo{person}{William~L Hamilton}, {and} \bibinfo{person}{Jure Leskovec}.}
  \bibinfo{year}{2018}\natexlab{}.
\newblock \showarticletitle{Graph convolutional neural networks for web-scale
  recommender systems}. In \bibinfo{booktitle}{\emph{Proceedings of the 24th
  ACM SIGKDD International Conference on Knowledge Discovery \& Data Mining}}.
  \bibinfo{pages}{974--983}.
\newblock


\bibitem[\protect\citeauthoryear{You, Ying, and Leskovec}{You
  et~al\mbox{.}}{2020}]%
        {you2020design}
\bibfield{author}{\bibinfo{person}{Jiaxuan You}, \bibinfo{person}{Zhitao Ying},
  {and} \bibinfo{person}{Jure Leskovec}.} \bibinfo{year}{2020}\natexlab{}.
\newblock \showarticletitle{Design space for graph neural networks}.
\newblock \bibinfo{journal}{\emph{Advances in Neural Information Processing
  Systems}}  \bibinfo{volume}{33} (\bibinfo{year}{2020}).
\newblock


\bibitem[\protect\citeauthoryear{Yu and Koltun}{Yu and Koltun}{2015}]%
        {yu2015multi}
\bibfield{author}{\bibinfo{person}{Fisher Yu} {and} \bibinfo{person}{Vladlen
  Koltun}.} \bibinfo{year}{2015}\natexlab{}.
\newblock \showarticletitle{Multi-scale context aggregation by dilated
  convolutions}.
\newblock \bibinfo{journal}{\emph{arXiv preprint arXiv:1511.07122}}
  (\bibinfo{year}{2015}).
\newblock


\bibitem[\protect\citeauthoryear{Yu, Lin, Liu, Ge, Ou, and Qin}{Yu
  et~al\mbox{.}}{2021}]%
        {yu2021self}
\bibfield{author}{\bibinfo{person}{Wenhui Yu}, \bibinfo{person}{Xiao Lin},
  \bibinfo{person}{Jinfei Liu}, \bibinfo{person}{Junfeng Ge},
  \bibinfo{person}{Wenwu Ou}, {and} \bibinfo{person}{Zheng Qin}.}
  \bibinfo{year}{2021}\natexlab{}.
\newblock \showarticletitle{Self-propagation Graph Neural Network for
  Recommendation}.
\newblock \bibinfo{journal}{\emph{IEEE Transactions on Knowledge and Data
  Engineering}} (\bibinfo{year}{2021}).
\newblock


\bibitem[\protect\citeauthoryear{Zeng, Zhang, Xia, Srivastava, Malevich,
  Kannan, Prasanna, Jin, and Chen}{Zeng et~al\mbox{.}}{2020}]%
        {zeng2020deep}
\bibfield{author}{\bibinfo{person}{Hanqing Zeng}, \bibinfo{person}{Muhan
  Zhang}, \bibinfo{person}{Yinglong Xia}, \bibinfo{person}{Ajitesh Srivastava},
  \bibinfo{person}{Andrey Malevich}, \bibinfo{person}{Rajgopal Kannan},
  \bibinfo{person}{Viktor Prasanna}, \bibinfo{person}{Long Jin}, {and}
  \bibinfo{person}{Ren Chen}.} \bibinfo{year}{2020}\natexlab{}.
\newblock \showarticletitle{Deep Graph Neural Networks with Shallow Subgraph
  Samplers}.
\newblock \bibinfo{journal}{\emph{arXiv preprint arXiv:2012.01380}}
  (\bibinfo{year}{2020}).
\newblock


\bibitem[\protect\citeauthoryear{Zhang and Chen}{Zhang and Chen}{2018}]%
        {zhang2018link}
\bibfield{author}{\bibinfo{person}{Muhan Zhang} {and} \bibinfo{person}{Yixin
  Chen}.} \bibinfo{year}{2018}\natexlab{}.
\newblock \showarticletitle{Link prediction based on graph neural networks}.
\newblock \bibinfo{journal}{\emph{Advances in Neural Information Processing
  Systems}}  \bibinfo{volume}{31} (\bibinfo{year}{2018}),
  \bibinfo{pages}{5165--5175}.
\newblock


\bibitem[\protect\citeauthoryear{Zhang, Cui, Neumann, and Chen}{Zhang
  et~al\mbox{.}}{2018}]%
        {zhang2018end}
\bibfield{author}{\bibinfo{person}{Muhan Zhang}, \bibinfo{person}{Zhicheng
  Cui}, \bibinfo{person}{Marion Neumann}, {and} \bibinfo{person}{Yixin Chen}.}
  \bibinfo{year}{2018}\natexlab{}.
\newblock \showarticletitle{An end-to-end deep learning architecture for graph
  classification}. In \bibinfo{booktitle}{\emph{Thirty-Second AAAI Conference
  on Artificial Intelligence}}.
\newblock


\bibitem[\protect\citeauthoryear{Zhang, Li, Xia, Wang, and Jin}{Zhang
  et~al\mbox{.}}{2020}]%
        {zhang2020revisiting}
\bibfield{author}{\bibinfo{person}{Muhan Zhang}, \bibinfo{person}{Pan Li},
  \bibinfo{person}{Yinglong Xia}, \bibinfo{person}{Kai Wang}, {and}
  \bibinfo{person}{Long Jin}.} \bibinfo{year}{2020}\natexlab{}.
\newblock \showarticletitle{Revisiting graph neural networks for link
  prediction}.
\newblock \bibinfo{journal}{\emph{arXiv preprint arXiv:2010.16103}}
  (\bibinfo{year}{2020}).
\newblock


\bibitem[\protect\citeauthoryear{Zhang, Chan, and Jaitly}{Zhang
  et~al\mbox{.}}{2017}]%
        {zhang2017very}
\bibfield{author}{\bibinfo{person}{Yu Zhang}, \bibinfo{person}{William Chan},
  {and} \bibinfo{person}{Navdeep Jaitly}.} \bibinfo{year}{2017}\natexlab{}.
\newblock \showarticletitle{Very deep convolutional networks for end-to-end
  speech recognition}. In \bibinfo{booktitle}{\emph{2017 IEEE international
  conference on acoustics, speech and signal processing (ICASSP)}}. IEEE,
  \bibinfo{pages}{4845--4849}.
\newblock


\end{thebibliography}
\citestyle{acmauthoryear}

\end{document}